\documentclass[a4paper,feqn]{cas-dc}

\usepackage[authoryear,longnamesfirst]{natbib}

\def\tsc#1{\csdef{#1}{\textsc{\lowercase{#1}}\xspace}}
\tsc{WGM}
\tsc{QE}

\begin{document}
\let\WriteBookmarks\relax
\def\floatpagepagefraction{1}
\def\textpagefraction{.001}

\shorttitle{}    

\shortauthors{}  

\title [mode = title]{Redirecting the Flow: Image Customization through Attention Distribution Shift}  

\tnotemark[1] 


%

\author[1,2]{Jie Li}
\ead{lijie2002@smail.nju.edu.cn}

\author[1,3]{Suorong Yang}
\cormark[1]
\ead{sryang@smail.nju.edu.cn}

\author[4]{Jian Zhao}
\ead{jianzhao@nju.edu.cn}

\author[1,2]{Furao Shen}[orcid=0000-0002-7285-326X]
\cormark[1]
\ead{frshen@nju.edu.cn}

\affiliation[1]{
    organization={State Key Laboratory for Novel Software Technology, Nanjing University},
    city={Nanjing},
    country={China}
}

\affiliation[2]{
    organization={School of Artificial Intelligence, Nanjing University},
    city={Nanjing},
    country={China}
}

\affiliation[3]{
    organization={School of Computer Science, Nanjing University},
    city={Nanjing},
    country={China}
}

\affiliation[4]{
    organization={School of Electronic Science and Engineering, Nanjing University},
    city={Nanjing},
    country={China}
}

\cortext[1]{Corresponding author}



\begin{abstract}
Subject-driven image customization aims to generate images that not only follow textual instructions but also preserve the identity of a given reference subject. Existing approaches, including test-time fine-tuning, encoder-based methods, and token competition in shared attention spaces, suffer from limited efficiency, misalignment between extracted reference features and the generative process, and interference from irrelevant information. To address these limitations, we formulate the customization task as a distribution shift induced by incorporating reference images into text-to-image generation, and derive a Conditional Attention Distribution Shift formulation grounded in maximum entropy theory. Building on this formulation, we propose CustomShift, a dual-branch architecture based on Stable Diffusion 3. The Reference-Alignment Branch leverages self-attention between reference images and subject names to achieve layer-wise alignment with latent representations, while the Cross-Guidance Branch integrates textual and reference cues to guide generation. Experiments on the DreamBooth and Custom101 benchmarks demonstrate that our method consistently outperforms state-of-the-art approaches, achieving a better balance between semantic fidelity and subject consistency.
\end{abstract}




\begin{keywords}
Subject-Driven Image Customization \sep Flow Matching \sep Attention Distribution \sep Tuning-Free Generation
\end{keywords}

\maketitle

\section{Introduction}

\begin{figure*}
  \includegraphics[width=\textwidth]{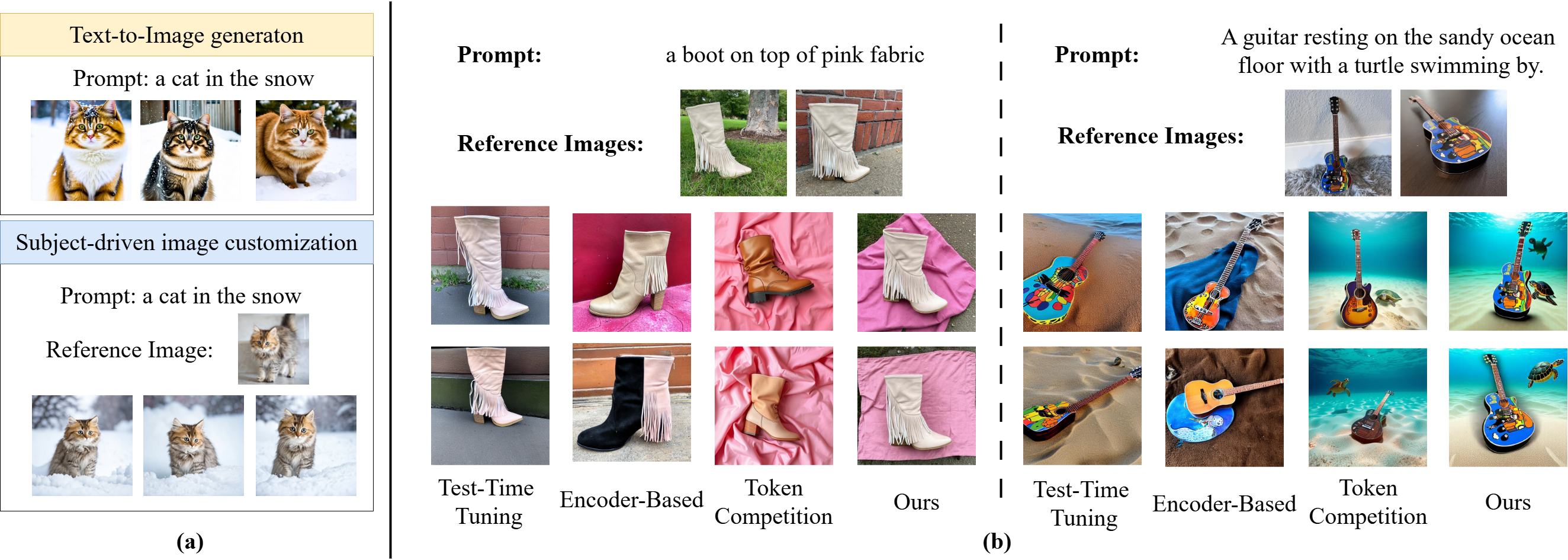}
  \caption{(a) Illustration of text-to-image generation versus subject-driven image customization, where the latter additionally uses reference images to preserve subject identity. (b) Qualitative comparison under subject-driven customization, showing that our method better balances instruction following and subject consistency compared to existing approaches.}
  \label{fig:intro}
\end{figure*}

Rapid advancements in text-to-image synthesis have significantly improved the realism and visual fidelity of generated images. However, as illustrated in Fig.~\ref{fig:intro} (a), although text-to-image models can produce semantically coherent outputs, they fail to maintain consistency of the generated subject. In professional applications such as comic illustration, advertising, and character design, maintaining consistent subject identity across diverse contexts is essential. Textual prompts alone are insufficient to capture the nuanced and distinctive details of specific entities, such as the intricate fur patterns of a cat, the delicate ornaments on a shoe, or the unique motifs on a guitar. By incorporating reference images, models can bypass the limitations of linguistic descriptions and directly ground the generation in the subject’s precise visual identity. This paradigm, commonly referred to as subject-driven image customization, leverages a small set of reference images together with textual instructions. It enables models to synthesize the target subject in novel scenes while maintaining both semantic fidelity and subject consistency, providing a robust framework for personalized visual content creation.

Current approaches typically address this task using pretrained text-to-image diffusion models \cite{ddpm,ldm,mmdit}. Since these models already possess strong instruction-following capabilities, the key challenge in customized generation lies in how to effectively incorporate image conditions.
As illustrated in Fig.~\ref{fig:intro}(b), existing methods can be broadly categorized into three groups: test-time tuning \cite{dreambooth,mcc,text_inversion}, encoder-based \cite{ssr,fastcomposer}, and token competition \cite{ominicontrol,omnigen,syncd} approaches.
Test-time tuning methods adapt the model to a specific subject by fine-tuning model parameters \cite{dreambooth,mcc} or learning a personalized token representation \cite{text_inversion,dreammatcher}. Such approaches require subject-specific optimization for each new subject, which limits their timeliness, and are highly sensitive to the number of reference images. In addition, they tend to overfit, often degrading instruction-following capabilities.
Encoder-based methods leverage dedicated image encoders to extract reference features and inject them into the generative model \cite{ssr,fastcomposer}. However, these externally extracted features are not dynamically aligned with the intermediate representations of the diffusion process, making it difficult for the model to utilize the reference information fully during generation.
Token competition methods \cite{syncd,ominicontrol,omnigen} incorporate reference images, textual prompts, and noised image tokens into a shared attention space through simple concatenation. All tokens are required to compete and interact within this shared attention space, leading to implicit fusion among heterogeneous conditions during attention allocation. 
As shown in Fig.~\ref{fig:intro}(b), this design can result in the inclusion of irrelevant background information or the weakening of textual guidance, ultimately reducing the stability and controllability of the generated results.

To overcome these challenges, we introduce the Conditional Attention Distribution Shift formulation. While standard text-to-image flow matching \cite{fm,mmdit,flux} achieves conditioning through joint self-attention on concatenated latents and prompts, subject-driven tasks require incorporating reference images as higher-order conditions. We treat these reference signals as a targeted distribution shift applied to the base generative path. By adopting a maximum-entropy objective, we derive routing weights that balance the expected compatibility of candidate sources with an entropy regularization term. This approach ensures a smooth transition between distributions, ultimately manifesting as a Softmax-governed integration of image, prompt, and reference signals.
Building on this formulation, we construct a dual-branch model, CustomShift. The reference alignment branch extracts consistent subject representations from reference images, while the Cross-Guidance Branch injects these signals into the generation process, facilitating controllable and stable subject-driven customization.
CustomShift avoids expensive test-time optimization by performing lightweight Low-Rank Adaptation (LoRA) \cite{lora} fine-tuning only during training, which significantly improves efficiency compared with test-time tuning methods. Our model defines the task as a probability distribution shift without the need for an external image encoder. The problem of encoding misalignment is naturally eliminated. By decoupling alignment and guidance, CustomShift reduces interference among reference images and achieves more stable customized generation.

Our primary contributions are summarized as follows:
\begin{itemize}
    \item \textbf{Theoretical Formulation}: We propose treating reference images as distribution shift terms within the attention mechanism. Based on maximum entropy theory, we derive the Conditional Attention Distribution Shift formulation, providing a rigorous theoretical foundation for subject-driven image customization.
    \item \textbf{Architectural Innovation}: Leveraging this formulation, we design CustomShift, a dual-branch architecture comprising a Cross-Guidance Branch and a Reference-Alignment Branch. The former enables the latent image to effectively consume both reference and prompt information, while the latter ensures the alignment of reference features at the structural level.
    \item \textbf{State-of-the-Art Performance}: Experiments on the DreamBooth and Custom101 benchmarks demonstrate that CustomShift achieves superior overall performance in terms of prompt alignment and subject consistency compared to existing methods.
\end{itemize}

\section{Related Work}

\subsection{Diffusion Models}

Diffusion models are a class of generative models that progressively transform a simple random distribution into a target data distribution through a sequence of denoising steps \cite{ddpm,sde,fm,dalle,glide,ldm,sdxl,mmdit,flux}. Early models such as DDPM \cite{ddpm}, DALL·E \cite{dalle}, and GLIDE \cite{glide} perform both training and inference directly in the pixel space. Later works, including the Stable Diffusion series \cite{ldm,sdxl,mmdit} and FLUX \cite{flux}, instead encode images into a latent space using variational autoencoders (VAEs) \cite{vae,savae} and perform the diffusion process in this latent space. By compressing images into lower-resolution latent representations, VAEs significantly reduce the computational cost of training and inference.
Diffusion models have evolved from convolutional U-Net backbones \cite{unet}, widely used in earlier systems such as Stable Diffusion 1.x/2.x and Stable Diffusion XL, toward Transformer-based architectures \cite{transformer,vit}. This shift is primarily driven by the superior scalability and representation capacity of Transformers, which better capture long-range dependencies and support large-scale training. Consequently, recent diffusion models, such as Stable Diffusion 3.x \cite{mmdit} and FLUX \cite{flux}, increasingly adopt Vision Transformer-based designs.
Since diffusion models themselves do not inherently process textual inputs, they typically rely on external text encoders—such as CLIP \cite{clip}, T5 \cite{t5}, or Transformer-based encoders \cite{transformer}—to encode textual prompts, which are then injected into the diffusion model via mechanisms such as cross-attention. Image conditions can also be incorporated through several approaches, including concatenating the conditioning image with the noisy input \cite{sr3}, or introducing auxiliary modules such as ControlNet \cite{controlnet} and adapters \cite{ip_adapter}.

\subsection{Subject-Driven Customization}

Subject-driven customization aims to generate images that faithfully preserve the identity of a target subject specified by reference images while following a given textual prompt. This task requires jointly modeling textual prompts and reference images.
Different from image editing methods \cite{sdedit,instructpix2pix,imagic}, which typically perform localized modifications on an existing image, image customization aims to synthesize new images with a much higher degree of freedom by conditioning on a specific subject extracted from reference images. 
While prompts can be reliably encoded by pretrained text encoders, representing and integrating reference images remains the central challenge.
From this perspective, existing methods can be broadly categorized into three paradigms. First, test-time tuning methods adapt the diffusion model to a target subject by fine-tuning model parameters using a few reference images, as in DreamBooth \cite{dreambooth}, Perfusion \cite{perfusion}, CustomDiffusion \cite{mcc}, and SVDiff \cite{svdiff}. A related line instead optimizes a learnable token while keeping the model fixed, such as Textual Inversion \cite{text_inversion} and DreamMatcher \cite{dreammatcher}.
Second, encoder-based methods introduce external image encoders to extract visual features, which are then injected into the U-Net. For example, IP-Adapter \cite{ip_adapter} employs decoupled cross-attention, while Blip-Diffusion \cite{blip-diffusion} and SSR-Encoder \cite{ssr} incorporate dedicated visual encoders. ELITE \cite{elite} fuses image and text embeddings through concatenation, whereas FastComposer \cite{fastcomposer} and InstantBooth \cite{instantbooth} further refine feature-level fusion strategies.
Third, token competition approaches unify image and text representations by directly feeding visual tokens into the attention space, allowing different modalities to interact and compete within the same attention mechanism. Representative methods include SynCD \cite{syncd} and OmniControl \cite{ominicontrol}, as well as large-scale multimodal generative models such as OmniGen \cite{omnigen} and Qwen \cite{qwen}, which adopt a unified token-based architecture for diverse multimodal tasks.

\begin{figure*}
 \centering 
 \includegraphics[width=0.92\linewidth]{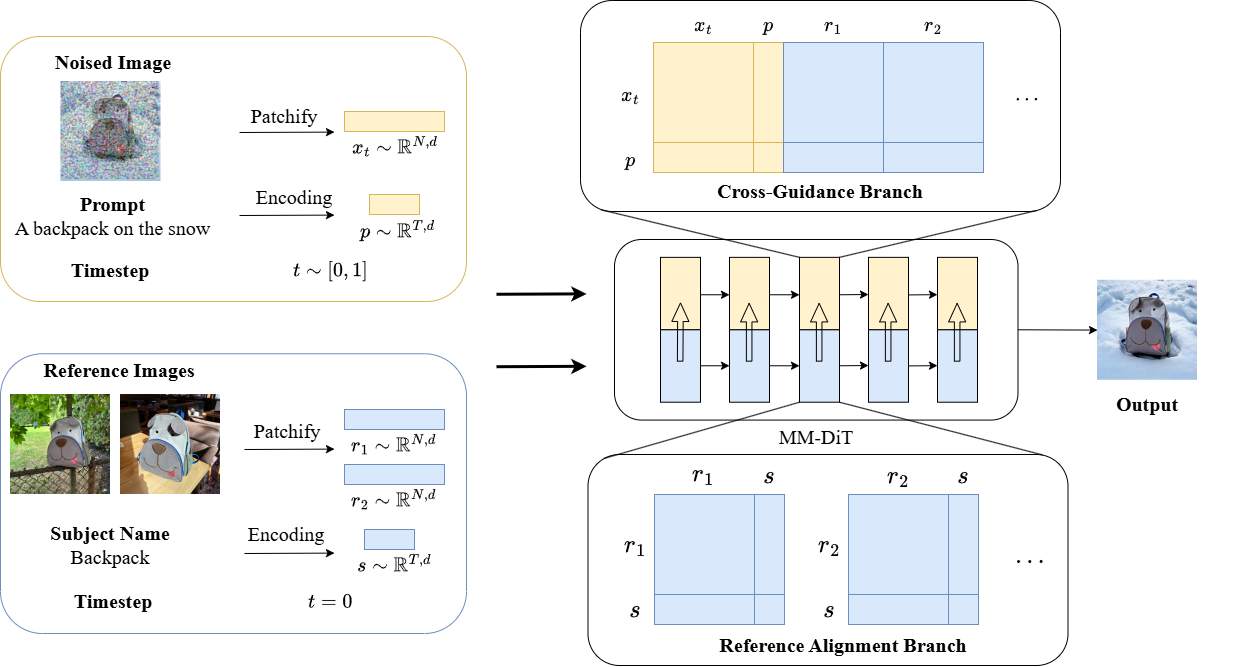}
\caption{CustomShift uses a dual-branch structure. (1) The \textit{Reference-Alignment Branch} leverages self-attention between reference images $r_i$ and the subject name $s$ to extract layer-wise, subject-aware features, which are then propagated in a unidirectional manner to the Cross-Guidance Branch, providing a robust foundation for stable and consistent customization. (2) Simultaneously, the \textit{Cross-Guidance Branch} modulates the latent $x_t$ and textual prompt $p$ by injecting visual priors from reference images, effectively aligning semantic content with the subject identity encoded in $R$.} 
 \label{fig:overview} 
 \end{figure*}

\section{Methodology}

\subsection{Preliminaries of Flow Matching}

We briefly introduce the basic concepts of flow matching. Let $p_0$ denote the data distribution and $p_1$ the prior distribution, typically chosen as a Gaussian. A trajectory ${x_t},{t\in[0,1]}$ describes the continuous evolution of a sample from $x_0 \sim p_0$ to $x_1 \sim p_1$. This evolution is governed by a vector field $v(x_t, t)$:
\begin{align}
\frac{d x_t}{dt} = v(x_t, t), \quad t \in [0,1],
\end{align}
which specifies the instantaneous velocity at state $x_t$ and time $t$. The collection of such trajectories defines a probability flow that transports samples from $p_0$ to $p_1$.
By modeling the generative flow as a time-dependent dynamical system, flow matching aims to reconstruct the underlying trajectories via a learned vector field, enabling precise prediction and transformation of probability distributions.

Following~\cite{fm}, we adopt a straight-line Optimal Transport (OT) interpolation to connect $x_0$ and $x_1$. :
\begin{align}
    x_t = (1-t)x_0 + t x_1,
\end{align}
Under this formulation, the corresponding target vector field is constant along the path:
\begin{align}
    \frac{d x_t}{dt} = x_1 - x_0,
\end{align}
Flow matching learns a neural network $v_\theta(x_t, t)$ to approximate the underlying vector field. During training, the model is optimized to match the ground-truth velocity along trajectories connecting $x_0 \sim p_0$ and $x_1 \sim p_1$. The training objective is defined as:
\begin{align}
\mathcal{L}_{FM} = \mathbb{E}_{t,x_0,x_1}
\|
v_\theta(x_t, t) - (x_1 - x_0)
\|^2.
\end{align}
At inference time, generation is performed by integrating the learned vector field starting from a sample $x_1 \sim p_1$ and evolving it backward along the flow to obtain $x_0$, which corresponds to a sample from the data distribution.

For conditional image generation tasks, such as text-to-image synthesis, the velocity network $v_\theta$ is conditioned on an additional input $\mathcal{C}$. This condition guides the predicted velocity field by steering the data trajectory toward the distribution specified by the condition. The training objective of conditional flow matching is given by:
\begin{align}
    \mathcal{L}_{CFM} = \mathbb{E}_{t,x_0,x_1} 
    \left\| 
    v_\theta(x_t, t \mid \mathcal{C}) - (x_1 - x_0)
    \right\|^2.
    \label{loss_cond}
\end{align}

In practice, models incorporate conditional information through architectural design. For example, Stable Diffusion 3 \cite{mmdit} with Multimodal Diffusion Transformer (MM-DiT) architecture first transforms the latent $x_t$ and the prompt $p$ into discrete patches and encodes them as token sequences. By concatenating these encoded tokens, the model applies a joint self-attention mechanism that allows for cross-modal communication. The output of this operation, denoted as $h$, serves as the refined hidden representation that propagates the integrated information to the subsequent Transformer block. Specifically, the image tokens and condition tokens are updated as:
\begin{align}
    &h_x=\mathrm{Attention}([x_t],[x_t;p],[x_t;p]), \nonumber \\
    &h_c=\mathrm{Attention}([p],[x_t;p],[x_t;p]), \label{eq:mmdit} \\
    &\mathrm{Attention}(q,k,v)=\mathrm{Softmax}\left(\frac{Q_qK_k^T}{\sqrt{d}}\right)V_v,
\end{align}
where $x_t$ and $p$ represent the patchified and encoded token sequences, $[x_t; p]$ denotes the concatenation of $x_t$ and $p$, and $d$ represents the dimensionality of the latent space. $Q_q$, $K_k$, and $V_v$ denote the corresponding linear projections for queries, keys, and values, respectively. The queries are computed from the image tokens and condition tokens separately, while the keys and values are shared across the concatenated token sequence. The query computes attention weights based on the concatenated key of $x_t$ and $p$, subsequently aggregating information from the corresponding concatenated value.

Compared to DDPM \cite{ddpm}, flow matching directly models continuous-time dynamics without predefined noise schedules, enabling more efficient training and faster inference with fewer sampling steps. As a result, recent generative models \cite{mmdit,flux,qwen} increasingly adopt flow-based formulations for better efficiency and scalability.

\subsection{Conditional Attention Distribution Shift}

The attention mechanism for $x_t$ and $p$ is symmetric in Eq.~\ref{eq:mmdit}. Therefore, we focus on the latent image sequence $x_t$, and denote it as $x$ in this section for simplicity. 
The image token attends to the image and prompt features as:
\begin{align}
    h_{x}^{base}
    & =
    \mathrm{Attention}([x],[x;p],[x;p]) \nonumber \\
    & =
    \mathrm{Softmax}
    \left(
    \frac{Q_x[K_x;K_p]^\top}{\sqrt d}
    \right)
    [V_x;V_p].
\end{align}
Define the attention scores and the normalized weights as:
\begin{align}
    S_x=\frac{Q_xK_x^\top}{\sqrt d},
    S_p=\frac{Q_xK_p^\top}{\sqrt d}, \nonumber \\
    [\rho_x,\rho_p]=\mathrm{Softmax}([S_x,S_p]).
\end{align}
The attention output is computed as a weighted sum of $V_x$ and $V_p$, with weights given by the distributions $\rho_x$ and $\rho_p$:
\begin{align}
    h_{x}^{base}=\rho_xV_x+\rho_pV_p.
\end{align}

In subject-driven image customization, the appearance of the generated subject is further specified by a set of reference images $R$. 
To incorporate this information, we introduce reference tokens
$R=[r_1;\cdots;r_N]$
and compute an additional compatibility score $S_R=\frac{Q_xK_R^\top}{\sqrt d}$
which measures how well the current hidden-state query aligns with the reference features.
We consider $S_R$ as a shift term, and induce a distribution shift in the attention routing, integrating the reference images $R$ and redirecting the generative flow toward customized outputs. Specifically, our goal is to find the updated conditional distribution as follows:
\begin{align}
h_x^{cond}=\pi_x V_x + \pi_p V_p + \pi_R V_R.
\end{align}

\begin{figure*}[ht]
    \centering
    \includegraphics[width=0.83\linewidth]{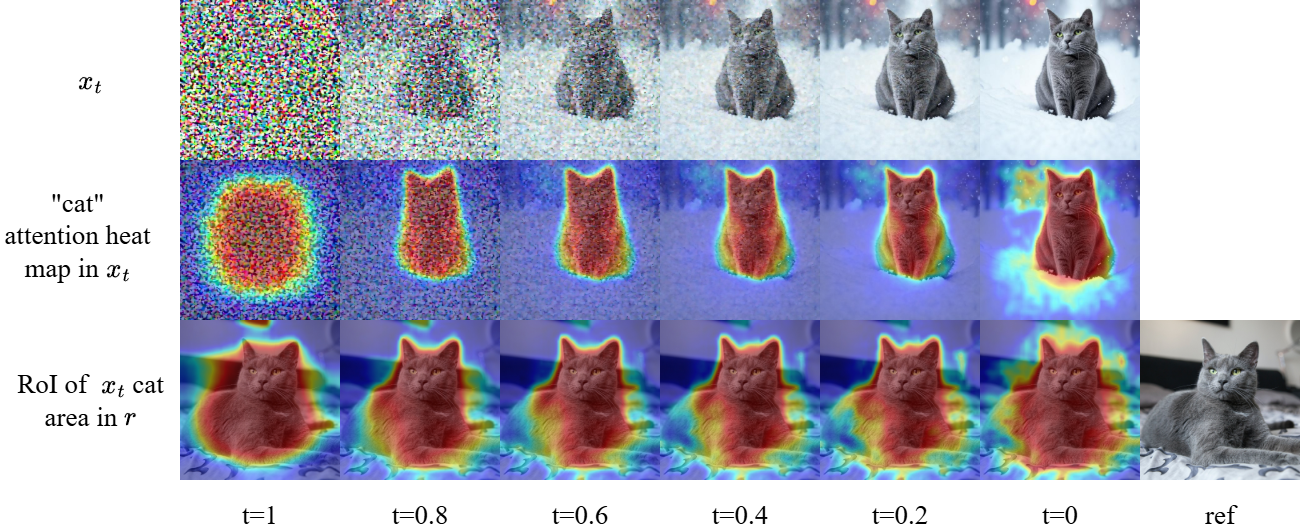}
    \caption{\textbf{Visualization of the Cross-Guidance Branch.} 
The top row shows the denoising process of $x_t$ across time steps $t$. 
The middle row displays the attention heatmap of the prompt token "cat" in $x_t$, reflecting the spatial localization of the subject. 
The bottom row illustrates the \textit{Region of Interest} (RoI) within the reference image $r$ that $x_t$ attends to, demonstrating that our model establishes accurate visual grounding from the earliest stages of diffusion ($t=1$).}
    \label{fig:heatmap}
\end{figure*}

Let $\mathcal{C}=\{x,p,R\}$ denote the set of candidate sources, corresponding to image features, prompt features, and reference features, respectively. 
A natural requirement is that the routing distribution should favor sources with higher compatibility scores. This can be captured by maximizing the expected score:
\begin{align}
    \sum_{c\in\mathcal C}\pi_c S_c,
\end{align}
However, maximizing this term alone would lead to a degenerate hard assignment that places all probability mass on the source with the largest score. To obtain a smooth and stable routing distribution, we further adopt the maximum entropy principle and define $\pi$ as the solution to the following entropy-regularized optimization:
\begin{align}
    & \max_{\pi}
    \sum_{c\in\mathcal C}\pi_c S_c
    + H(\pi), \nonumber \\
    & H(\pi)=-\sum_{c\in\mathcal C}\pi_c\log\pi_c, \nonumber \\
    & s.t. \quad \{\pi | \pi_c \ge 0, \Sigma_{c\in \mathcal{C}} \pi_c = 1\},
\end{align}
$H(\pi)$ is the Shannon entropy \cite{entropy}. The first term encourages routing toward sources with higher compatibility scores, while the entropy term prevents overly sharp assignments and preserves uncertainty among competing sources.
Introducing a Lagrange multiplier $\lambda$ for the normalization constraint, we obtain:
\begin{align}
    \mathcal{L}(\pi,\lambda)
    =
    \sum_{c\in\mathcal C}\pi_c S_c
    -\sum_{c\in\mathcal C}\pi_c\log\pi_c
    +\lambda\left(\sum_{c\in\mathcal C}\pi_c-1\right),
\end{align}
Taking the derivative with respect to $\pi_c$ and setting it to zero gives:
\begin{align}
    \frac{\partial \mathcal L}{\partial \pi_c}
    =
    S_c-(\log\pi_c+1)+\lambda=0, \nonumber \\
    \pi_c = \exp (S_c + \lambda - 1) =Z \cdot \exp(S_c),
\end{align}
where $Z$ is a $\lambda$ related constant. The probability normalization constraint $\sum_{c\in\mathcal C}\pi_c=1$ gives the distribution:
\begin{align}
    Z=\exp & (\lambda-1)=\frac{1}{\sum_{c'\in\mathcal C} \exp (S_{c'})}, \nonumber \\
    \pi&_c^*=\frac{\exp(S_c)}{\sum_{c'\in\mathcal C}\exp(S_{c'})}.
\end{align}
Therefore, the reference-conditioned routing distribution can be viewed as a distribution shift:
\begin{align}
    [\pi_x,\pi_p,\pi_R]
    &=
    \mathrm{Softmax}([S_x,S_p,S_R]), \\
    h_x^{cond} &= \mathrm{Softmax}([S_x,S_p,S_R])[V_x;V_p,V_R] \nonumber \\ &= \mathrm{Attention}([x],[x;p;R],[x;p;R]).
    \label{h_cond}
\end{align}
Similarly, the hidden state of the prompt $p$ at the next layer can be expressed as:
\begin{align}
    h_p^{cond}=\mathrm{Attention}([p],[x;p;R],[x;p;R]).
\end{align}

\subsection{CustomShift}

We develop CustomShift to realize the Conditional Attention Distribution Shift formulation introduced in Eq.~\ref{h_cond}. Beyond the joint self-attention in Eq.~\ref{eq:mmdit}, both $x_t$ and $p$ directly extract subject-related information from all reference images via cross-attention. At the same time, each reference image is processed independently, without being influenced by the noisy latent $x_t$ or the prompt $p$, ensuring that each reference provides a clean and reliable visual signal for subject guidance.
This design differs from prior token competition approaches that permit unrestricted interactions among image and text tokens (e.g., \cite{syncd, ominicontrol}), which can introduce interference between reference concepts.
Based on this principle, CustomShift adopts a dual-branch architecture consisting of a Reference-Alignment Branch, which ensures faithful subject representation and stable generation, and a Cross-Guidance Branch, responsible for subject extraction and reference consistency. An overview of CustomShift is illustrated in Fig.~\ref{fig:overview}.

\begin{figure}
  \centering
  \includegraphics[width=\linewidth]{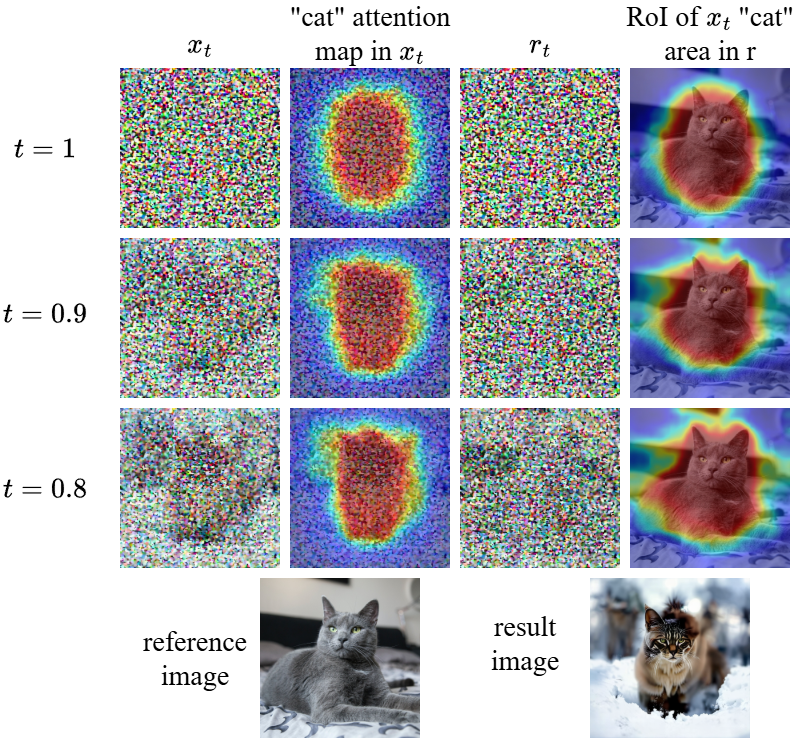}
  \caption{Impact of synchronized noise injection on reference images. Early-stage subject localization and structural grounding are severely compromised due to high noise levels, resulting in structural misalignment in the final generation.}
  \label{fig:ref_time}
\end{figure}

\begin{table*}[t]
\centering
\caption{Comparison with state-of-the-art methods on DreamBooth and Custom101 benchmarks.}
\begin{tabular}{lcccc|cccc}
    \toprule
    & \multicolumn{4}{c}{DreamBooth} & \multicolumn{4}{c}{Custom101} \\
    \cmidrule(lr){2-5} \cmidrule(lr){6-9}
    Method 
    & CLIP$_{Text}$ 
    & CLIP$_{Image}$ 
    & DINO$_{Image}$ 
    & Average 
    & CLIP$_{Text}$ 
    & CLIP$_{Image}$ 
    & DINO$_{Image}$ 
    & Average \\
    \midrule
    DreamBooth
    & 0.7603 & \underline{0.7856} & 0.6144 & 0.7201
    & 0.7339 & 0.7495 & 0.5308 & 0.6714 \\
    
    CustomDiffusion
    & 0.7803 & 0.7686 & 0.6125 & 0.7205
    & 0.7412 & 0.7183 & 0.4918 & 0.6504 \\
    
    SSR Encoder
    & 0.7300 & \textbf{0.7886} & \underline{0.6284} & 0.7157
    & 0.6855 & \textbf{0.7685} & 0.5710 & 0.6750 \\
    
    OminiControl
    & 0.7695 & 0.7590 & 0.5622 & 0.6969
    & 0.7402 & 0.7163 & 0.4992 & 0.6519 \\
    
    OmniGen
    & 0.7720 & 0.7605 & 0.6152 & 0.7159
    & 0.7598 & 0.7067 & 0.5089 & 0.6585 \\
    
    SynCD 
    & 0.7637 & 0.7744 & 0.6129 & 0.7170
    & 0.7017 & 0.7432 & 0.5706 & 0.6718 \\

    \midrule

    CustomShift (1shot) 
    & \textbf{0.7866} & 0.7622 & 0.6255 & \underline{0.7248} 
    & \textbf{0.7622} & 0.7383 & \underline{0.5844} & \underline{0.6950}  \\
    
    CustomShift 
    & \underline{0.7813} & 0.7784 & \textbf{0.6712} & \textbf{0.7436}
    & \underline{0.7603} & \underline{0.7554} & \textbf{0.6135} & \textbf{0.7097} \\
    \bottomrule
\end{tabular}
\label{tab:main_results}
\end{table*}

\subsubsection{Reference-Alignment Branch.}
In Eq.~\ref{h_cond}, the noisy image $x_t$ needs to obtain subject-related features from the reference set $R$ through an attention mechanism. Therefore, it is necessary to construct reference representations that are aligned with $x_t$.
In flow matching diffusion models, the generation process involves two dimensions: the time dimension $t \sim [0,1]$ and the model depth from shallow to deep layers. As time decreases from $1$ to $0$, the noise level of $x_t$ gradually reduces. Meanwhile, deeper layers typically encode more abstract semantic representations.
We do not incorporate time information into the reference images, as aligning them with $x_t$ along the time dimension introduces significant noise and hinders effective guidance. As shown in Fig.~\ref{fig:ref_time}, when time is applied to the reference images, they become heavily corrupted by noise at early diffusion stages (large $t$), making it difficult for $x_t$ to locate and extract meaningful subject information. Consequently, during the critical stage where $x_t$ establishes the overall layout and structure of the subject, the reference fails to provide useful guidance, leading to a misinterpretation of the target subject. In contrast, as illustrated in Fig.~\ref{fig:heatmap}, when the reference images are kept free of time-dependent noise, $x_t$ can effectively attend to and extract relevant subject information from the reference even at early stages of the diffusion process. This enables the model to localize the subject and incorporate its visual characteristics during the initial layout construction. Therefore, instead of enforcing alignment along the time dimension, we set timestep $t=0$ in this branch and focus on aligning reference information with $x_t$ at the model layer level, where their representations are more semantically compatible.
As illustrated in Fig. \ref{fig:overview}, the Reference-Alignment Branch processes each reference image in conjunction with its corresponding prompt $s$, which identifies the target subject for customization. By performing self-attention conditioned on $s$, the reference tokens facilitate the reconstruction of the subject representation as dictated by the textual instruction.
In practice, we achieve this by placing the reference images and the generated image in the same batch and feeding them through the transformer blocks simultaneously, so that their features at each layer share the same level of abstraction.

\subsubsection{Cross-Guidance Branch.}
The Cross-Guidance Branch performs the conditional attention in Eq. \ref{h_cond}, allowing the noised image tokens $x_t$ and prompt tokens $p$ to attend to $x_t$, $p$, and the reference tokens $R$. Tokens capture intrinsic structure through self-attention over their own features, and are updated based on cross-attention with information from other sources.
As illustrated in Fig.~\ref{fig:heatmap}, the model primarily establishes the object layout during the early stages of diffusion (before $t=0.8$). In this phase, the queries from $x_t$ obtain coarse shape information of the subject from $p$, while leveraging self-attention to localize the subject and construct its basic structure. At this early inference stage, the model is already able to identify the relevant regions of the subject in the reference images, and subsequently incorporates features from $R$ to capture and integrate the customized subject characteristics. We will discuss the form of $R$ in the next section.
As the inference process proceeds, the model progressively refines the spatial structure and appearance of the subject by continuously retrieving visual evidence from the reference branch. 
Eventually, the generated representation converges to a subject configuration that is consistent with the target specified by the prompt and the reference images.

During training, we focus only on the flow matching objective of the noised image without considering reconstruction of the reference images, and therefore adopt the same loss function as in Eq.~\ref{loss_cond}, where $c=\{p,R,s\}$. Inspired by \cite{syncd}, we employ normalized image and text guidance vectors during the inference stage. Given our dual-branch architecture, we treat the shift parameter $s$ from the Reference-Alignment Branch as a fundamental condition and refrain from nullifying it during unconditional estimation. The composite guidance is formulated as follows:
\begin{align}
    \hat{v}_{\theta}(x_t,t|p,R,s) &= v_{\theta}(x_t,t|\varnothing,\varnothing,s) + \lambda_R \frac{\|g\|}{\|g_R\|} g_R + \lambda_p \frac{\|g\|}{\|g_p\|} g_p, \nonumber \\
    g_R &= v_{\theta}(x_t,t|\varnothing,R,s) - v_{\theta}(x_t,t|\varnothing,\varnothing,s), \nonumber \\
    g_p &= v_{\theta}(x_t,t|p,R,s) - v_{\theta}(x_t,t|\varnothing,R,s), \nonumber \\
    \|g\| &= \min(\|g_p\|, \|g_R\|).
    \label{eq:infer}
\end{align}

\begin{figure*}[h]
  \centering
  \includegraphics[width=0.97\linewidth]{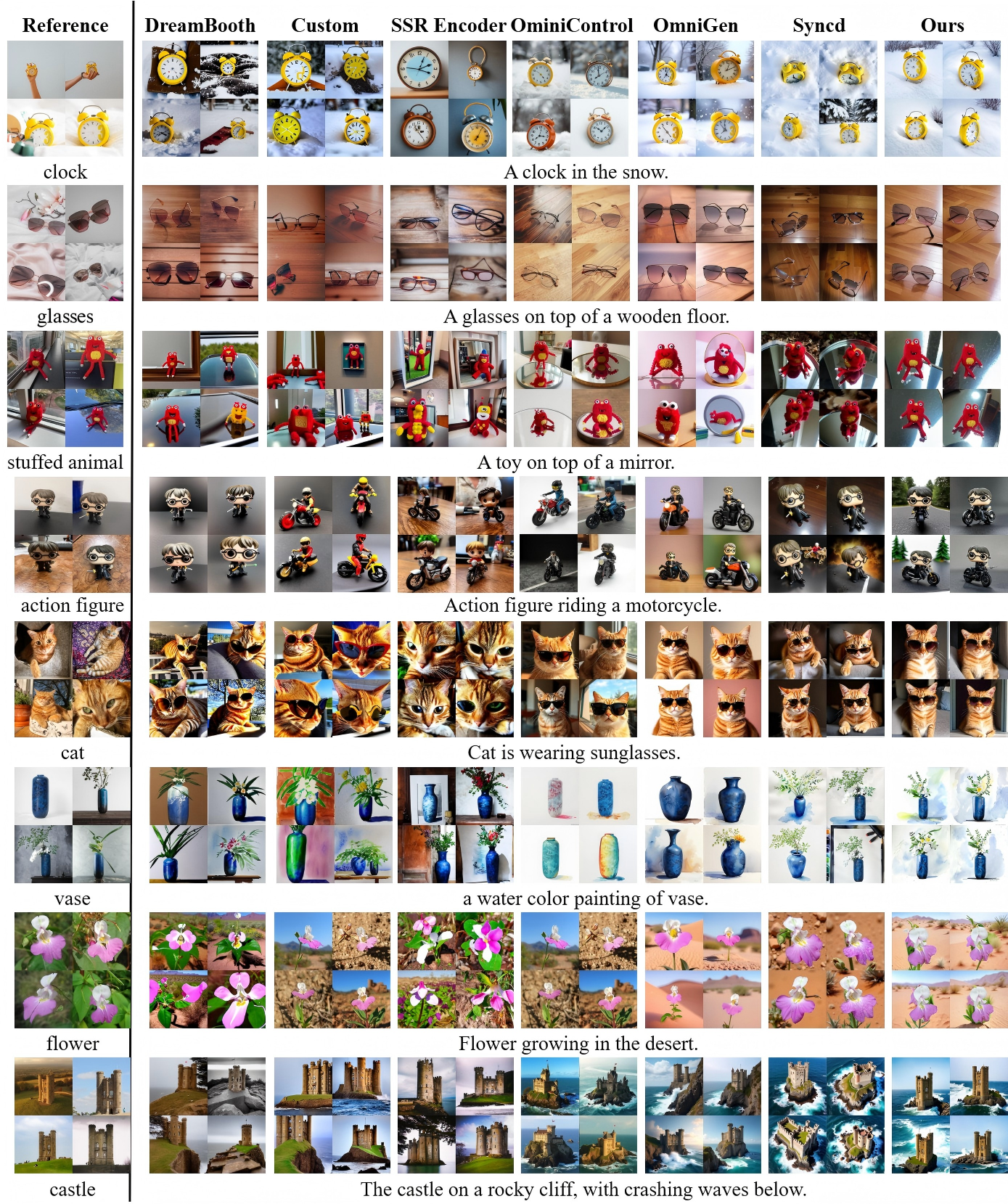}
  \caption{Qualitative comparison of subject-driven image generation. The first column presents the reference images, followed by results from several baseline methods, with our proposed method shown in the final column. Overall, our approach demonstrates superior performance in both prompt-following capability and identity fidelity relative to the reference subjects.}
  \label{fig:qualitative}
\end{figure*}

\section{Experiments}

\subsection{Implementation Details}

We fine-tune our model on top of Stable Diffusion 3 \cite{mmdit} using LoRA \cite{lora}. The model is trained on the SynCD dataset \cite{syncd}, which contains more than 90,000 groups of data, each consisting of 2–3 prompt–image pairs. We use DeepSeek \cite{deepseek} to extract the subject name $s$ from each prompt as the input to the Reference-Alignment Branch. For each group, one prompt–image pair is selected as the target sample, while the remaining images are used as reference images. No additional fine-tuning is required at inference time. We set $\lambda_R=\lambda_p=7$ in Eq. \ref{eq:infer}.

\subsection{Comparison with Previous Methods}

Our comparison methods include training-time tuning approaches such as DreamBooth \cite{dreambooth} and Custom Diffusion \cite{mcc}, encoder–based method SSR-Encoder \cite{ssr}, and token competition methods including OmniControl \cite{ominicontrol}, OmniGen \cite{omnigen}, and SynCD \cite{syncd}. We reproduce all methods using their officially released codebases and pretrained weights.
We evaluate all methods on the DreamBooth \cite{dreambooth} and Custom101 \cite{mcc} benchmarks. DreamBooth consists of 30 subject categories, each containing 5--6 reference images. In contrast, Custom101 is a larger benchmark comprising 101 subjects, with 3--15 reference images available for each subject.
For the number of reference images, we follow the configurations recommended in the original papers to ensure fair comparisons. Specifically, DreamBooth and CustomDiffusion, as tuning-based methods, utilize all available reference images during adaptation. SSR Encoder, OmniControl, and OmniGen are evaluated in the 1-shot setting, whereas SynCD adopts a 3-shot setting. By default, CustomShift uses 3 reference images, consistent with SynCD, and we additionally report 1-shot results to facilitate direct comparison with 1-shot approaches.

\subsubsection{Quantitative Results}

We use CLIP$_{Text}$ \cite{clip} to evaluate the similarity between the generated images and the prompts, which reflects the model’s instruction-following ability. CLIP$_{Image}$ \cite{clip} and DINO$_{Image}$ \cite{dino} are used to measure the similarity between the generated images and the reference images. The former focuses on semantic consistency, while the latter emphasizes similarity in visual appearance. We adopt CLIP ViT-B/32 and DINO ViT-S/16 for evaluation. Due to the multi-objective nature of subject-driven image customization, we report the average of the three metrics as an overall performance indicator. All metrics are higher-is-better. The comparison results are shown in Table~\ref{tab:main_results}.

CustomShift establishes a new state-of-the-art on both the DreamBooth and Custom101 benchmarks, achieving the highest Average scores of 0.7436 and 0.7097, respectively. Notably, even under the 1-shot setting, CustomShift achieves Average scores of 0.7248 and 0.6950, ranking second among all compared methods on both benchmarks. Examining the individual metrics, our methods attain the best performance on both CLIP$_{Text}$ and DINO$_{Image}$, demonstrating its strong capability in preserving subject identity while maintaining faithful adherence to textual instructions.

Although its CLIP$_{Image}$ score is slightly lower than those of SSR Encoder and DreamBooth, these methods exhibit noticeably weaker performance on the remaining metrics. This difference can be attributed to the characteristics of their underlying designs. As an encoder-based approach, SSR Encoder is particularly effective at capturing high-level semantic appearance features emphasized by CLIP$_{Image}$, such as overall visual similarity, color distribution, and coarse object characteristics. However, it is less effective at preserving fine-grained instance-specific details and structural cues that are crucial for DINO$_{Image}$, while also lacking strong text-conditioned controllability. DreamBooth and CustomDiffusion are both based on test-time tuning. DreamBooth performs full-parameter fine-tuning on the reference images, which tends to overfit the model to the visual appearance of the reference subject, resulting in strong image-similarity scores but relatively weaker text-image alignment. In contrast, CustomDiffusion updates only the cross-attention layers, reducing the risk of overfitting. However, the limited adaptation capacity prevents it from fully capturing subject-specific information, leading to inferior image-similarity and identity-preservation performance.

For token-competition-based methods such as OmniGen, OmniControl, and SynCD, image tokens and prompt tokens compete within a shared attention space. Such competition introduces interference between textual and visual conditions, making it difficult for the model to simultaneously preserve subject identity and follow textual instructions. Consequently, these methods generally achieve lower CLIP$_{Text}$ and image-similarity scores. In contrast, by decoupling reference alignment from generation guidance, CustomShift avoids direct competition between reference and prompt information. As a result, it retains highly competitive CLIP$_{Image}$ scores while achieving the highest CLIP$_{Text}$ and DINO$_{Image}$ performance. Furthermore, the strong results of CustomShift (1-shot) indicate that the proposed architecture remains highly effective even with only a single reference image, demonstrating the robustness and data efficiency of our design. Overall, these results show that our dual-branch architecture achieves a superior balance between textual controllability and reference-subject consistency, leading to the best overall performance across both benchmarks.

\subsubsection{Qualitative Results}

The qualitative results, illustrated in Fig.~\ref{fig:qualitative}, showcase the robustness of CustomShift across a diverse array of subject categories, ranging from organic entities (e.g., cats, stuffed animals) to rigid manufactured objects (e.g., clocks, vases). Our framework demonstrates superior versatility in handling various customization tasks, including seamless background replacement, complex object interactions, and artistic style transfers.
A detailed comparative analysis reveals significant limitations in existing methods regarding subject fidelity. Specifically, DreamBooth and SSR Encoder struggle to preserve the identity of the "glasses" "stuffed animal" and "castle", while CustomDiffusion fails to maintain the distinctive traits of the "clock" and "action figure". Even recent approaches like OmniGen and SynCD exhibit noticeable identity drifting on "castle" and "vase", failing to replicate the reference subject's core features faithfully.
Furthermore, several baselines demonstrate a clear deficiency in instruction-following capability. For instance, DreamBooth and SynCD fail to render an "action figure riding a motorcycle", while SSR-Encoder struggles with contextual modifications, such as placing a "clock" in a snowy background or adding "sunglasses" to a cat. We also observe failures in capturing complex scene logic: DreamBooth, CustomDiffusion, and OmniGen all fail to generate the "mirror" requested for the stuffed animal scene. Moreover, the "watercolor painting" style is inconsistently applied by DreamBooth, CustomDiffusion, and SSR-Encoder.
In contrast, CustomShift consistently succeeds where others fail. This demonstrates that our Conditional Attention Distribution Shift effectively guides the denoising process to strike a precise balance between global semantic instructions and local subject fidelity.

\subsection{Ablation Study}

\begin{table}[h]
    \centering
    \caption{Ablation study on the DreamBooth benchmark. We evaluate (1) different subject prompt settings in the Reference-Alignment Branch and (2) different reference interaction strategies for incorporating reference image information.}
    \resizebox{\columnwidth}{!}
    {\begin{tabular}{lcccc}
        \toprule
        Metrics & CLIP$_{Text}$ & CLIP$_{Image}$ & DINO$_{Image}$ & Average \\
        \midrule
        No prompt & 0.7788 & 0.7783 & 0.6704 & 0.7425 \\
        Generic prompt & 0.7788 & 0.7778 & 0.6714 & 0.7427 \\
        Wrong prompt (Spaceship) & 0.7793 & 0.7773 & 0.6678 & 0.7415 \\
        Wrong prompt (Cute) & 0.7808 & 0.7764 & 0.6656 & 0.7409 \\
        Token-competition & 0.7852 & 0.7446 & 0.5594 & 0.6964 \\
        CustomShift & 0.7813 & 0.7784 & 0.6712 & 0.7436 \\
        \bottomrule
    \end{tabular}}
    \label{tab:ablation}
\end{table}

\subsubsection{Importance of Subject Name Prompt $s$}

We evaluate the influence of subject prompt quality in the Reference-Alignment Branch on the DreamBooth benchmark. As shown in Table~\ref{tab:ablation}, using the specific subject prompt achieves the best overall performance. When the prompt is removed entirely (No prompt) or replaced with a generic description (Generic prompt: ``the foreground object''), the performance drops slightly across most metrics. We further evaluate two inaccurate prompt settings: a noun prompt (Spaceship) that never appears in the dataset and an adjective prompt (Cute) that does not specify the subject identity. Both settings lead to a noticeable decline in DINO$_{Image}$ and Average scores. These results indicate that the more accurately the prompt describes the target subject, the more effectively the Reference-Alignment Branch can identify and align subject-related features from the reference images.

Nevertheless, the degradation caused by imprecise prompts remains relatively limited. Even with generic or incorrect prompts, all variants still achieve Average scores above 0.740, outperforming the strongest baseline methods reported in Table~\ref{tab:main_results}. This observation demonstrates the robustness of our reference-alignment mechanism. While an accurate subject prompt further enhances alignment quality, the model is still capable of extracting meaningful subject-specific information from the reference images themselves, resulting in consistently strong customization performance.

\subsubsection{Effect of Reference Interaction Strategy}

We compare our independent alignment strategy with the token competition paradigm. Specifically, we replace the proposed independent alignment mechanism with a shared-attention design, where tokens from all reference images are jointly processed and allowed to compete for attention, rather than being aligned separately in isolated sequences. 
As reported in Table~\ref{tab:ablation}, enabling token competition leads to a noticeable performance degradation across most metrics. Although CLIP$_{Text}$ increases slightly, DINO$_{Image}$ drops to 0.5594 and the average score decreases to 0.6964. We attribute the gain in CLIP$_{Text}$ to the intensified competition between textual and reference tokens within a shared attention space, which shifts the model's focus toward textual conditions and thus improves text-image alignment. However, such competition weakens the contribution of subject-specific reference cues. Furthermore, the coupled processing of multiple references introduces inter-sample noise, including irrelevant background content, viewpoint discrepancies, and lighting variations, which contaminate the extracted subject representation. Consequently, identity preservation deteriorates despite improved prompt adherence. In contrast, our independent alignment strategy derives the Conditional Attention Distribution Shift from instance-specific features, improving both subject fidelity and semantic consistency.

\subsection{Effect of the Number of Reference Images}

\begin{figure}
    \centering
    \includegraphics[width=\linewidth]{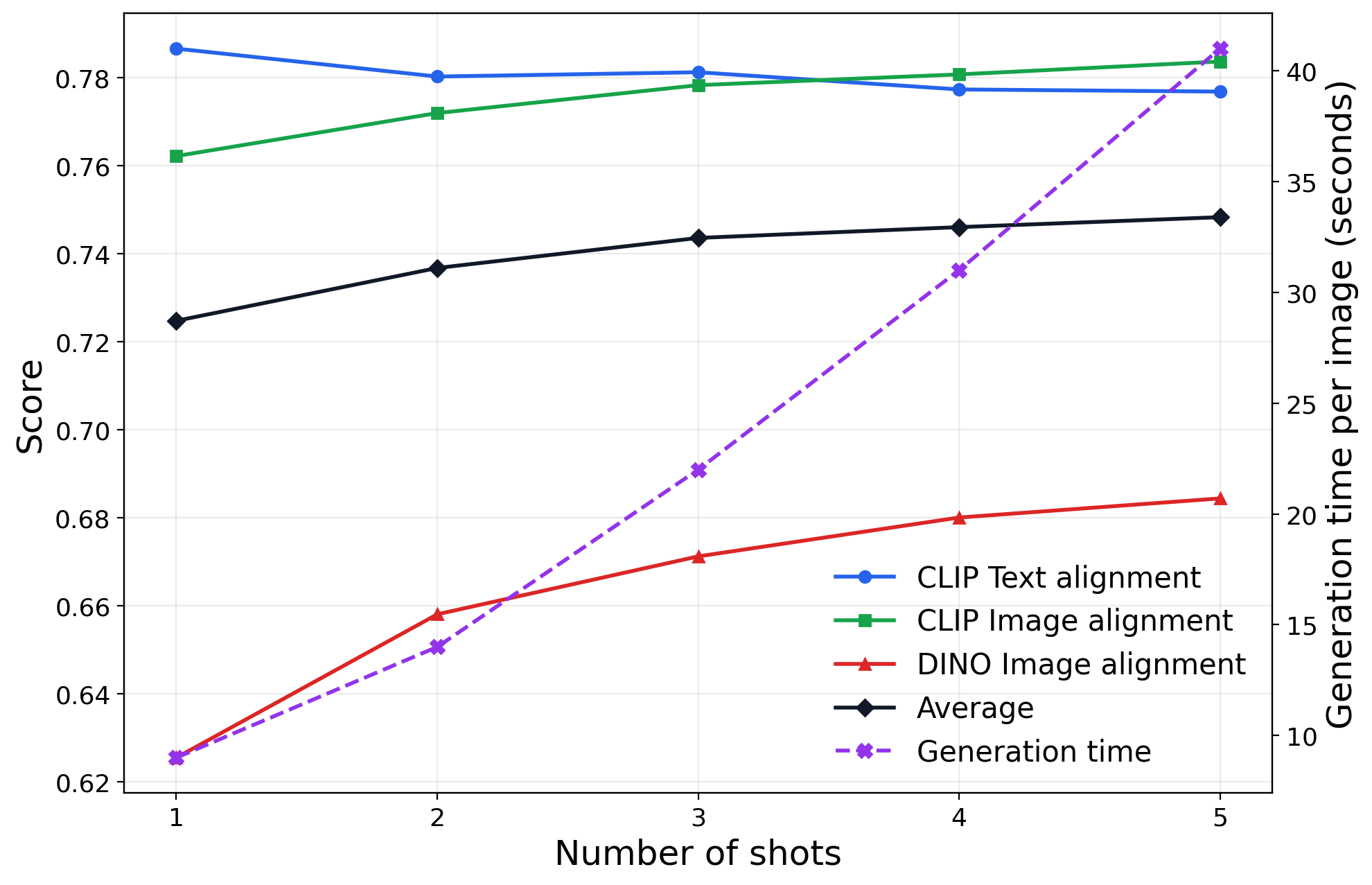}
    \caption{Effect of the number of reference images on DreamBooth.}
    \label{fig:few-shot}
\end{figure}

We evaluate the effect of the number of reference images on the DreamBooth benchmark, reporting both quantitative metrics and the inference time required to generate a single image. All timing measurements are conducted on a single NVIDIA RTX 4090 GPU. As shown in Fig.~\ref{fig:few-shot}, increasing the number of reference images consistently improves the Average score, indicating that additional references provide more comprehensive subject information. However, the performance gain gradually saturates as more reference images are introduced. Meanwhile, the inference time increases approximately linearly with the number of reference images.

In principle, according to Eq.~\ref{h_cond} and the proposed dual-branch architecture, the computational complexity of attention operations grows as $O(n^2)$ with respect to the number of reference tokens. Nevertheless, since our method operates in a few-shot setting where the number of reference images remains relatively small, the practical runtime growth is closer to linear. Despite this favorable scaling behavior, the computational cost increases faster than the performance gain, suggesting diminishing returns from adding more references.

Examining the individual metrics, both image-similarity metrics improve as more reference images are provided, with DINO$_{Image}$ exhibiting the most significant increase. This observation indicates that additional references help the model capture more diverse instance-specific appearance details and structural characteristics, thereby improving subject fidelity. In contrast, CLIP$_{Text}$ shows a slight decline as the number of reference images increases. We attribute this behavior to the stronger influence of visual guidance from multiple references, which encourages the model to focus more heavily on preserving subject identity and appearance consistency, occasionally at the expense of strict adherence to textual instructions. Overall, these results demonstrate a trade-off between subject fidelity and computational efficiency, with three reference images providing a favorable balance between performance and inference cost.

\section{Conclusion}
In this paper, we first introduce a principled formulation that interprets reference images as inducing a distribution shift within the attention mechanism, and derive the Conditional Attention Distribution Shift based on maximum entropy theory. Building on this formulation, we design a dual-branch architecture that decouples reference alignment and generation guidance, enabling more effective integration of reference and textual information. Extensive experiments on the DreamBooth and Custom101 benchmarks demonstrate that our method achieves state-of-the-art performance, significantly improving both prompt-following capability and subject fidelity.
Overall, CustomShift provides a theoretically grounded and empirically superior framework solution for customized image generation, highlighting the importance of structured attention modeling for balancing semantic consistency and subject preservation. 
We hope this work offers useful insights into handling multimodal inputs and contributes to the development of more robust and controllable generative models.

\bibliographystyle{cas-model2-names}

\bibliography{cas-refs}



\end{document}